# On anthropomorphic decision making in a model observer

Ali R. N. Avanaki[a], Kathryn S. Espig[a], Tom R. L. Kimpe[b], Andrew D. A. Maidment[c]
[a]Barco Healthcare, Beaverton, OR; [b]Barco Healthcare, Kortrijk, Belgium;
[c]University of Pennsylvania, Philadelphia, PA

## ABSTRACT

By analyzing human readers' performance in detecting small round lesions in simulated digital breast tomosynthesis background in a location known exactly scenario, we have developed a model observer that is a better predictor of human performance with different levels of background complexity (i.e., anatomical and quantum noise). Our analysis indicates that human observers perform a lesion detection task by combining a number of sub-decisions, each an indicator of the presence of a lesion in the image stack. This is in contrast to a channelized Hotelling observer, where the detection task is conducted holistically by thresholding a single decision variable, made from an optimally weighted linear combination of channels. However, it seems that the sub-par performance of human readers compared to the CHO cannot be fully explained by their reliance on sub-decisions, or perhaps we do not consider a sufficient number of sub-decisions. To bridge the gap between the performances of human readers and the model observer based upon sub-decisions, we use an additive noise model, the power of which is modulated with the level of background complexity. The proposed model observer better predicts the fast drop in human detection performance with background complexity.

## 1. INTRODUCTION

Validation of an imaging system is challenging due to the large number of system parameters that must be considered. Conventional methods involving clinical trials are limited by cost and duration, and in the instance of systems using ionizing radiation, the requirement for the repeated irradiation of volunteers. We are proponents of an alternative, in the form of Virtual Clinical Trials (VCTs) based on models of human anatomy, image acquisition, display and processing, and image analysis and interpretation. In this collaboration, researchers at the University of Pennsylvania develop anatomy and image simulation [7, 8, 9] and researchers at Barco develop display simulation and virtual observers. Barco has developed anthropomorphic model observers that predict typical human observers better than commonly used model observers, which are designed after ideal observers with some concessions for tractability.

Previously we reported [1, 2] that by embedding properties of human visual system (HVS) as pre-processing steps to a commonly used model observer (multi-slice channelized Hotelling observer – msCHO [3]), we can predict the performance of a human observer with respect to changes in viewing distance, browsing speed and display contrast when reading digital breast tomosynthesis (DBT) images. This is achieved by simulation of spatiotemporal contrast sensitivity function, psychometric function, as well as Barten's contrast masking [15] adapted to continuous browsing of digital breast tomosynthesis data.

In this paper, we explore an alternative to the existing decision block of a traditional observer pipeline. This alternative is designed based upon our analysis of the missed detection cases in our previous reader [4] and thus allows for better prediction of human observers' performance with increasing background complexity.

## 2. METHODS

Synthetic breast images are generated using the breast anatomy and imaging simulation pipeline shown in Fig. 1 [7, 16-19]. Normal breast anatomy is simulated with a recursive partitioning algorithm using octrees [7]. Lesions can be included automatically based upon a configurable set of rules [20]. Phantom deformation due to clinical breast positioning and compression is simulated using a finite element (FE) model and rapid post-FE software [21]. DBT image acquisition is simulated by ray tracing projections, assuming a polyenergetic x-ray beam without scatter, and an ideal detector model. Processed or reconstructed images are obtained using image reconstruction and processing software (Real-Time Tomography, LLC, Villanova, PA) [9].

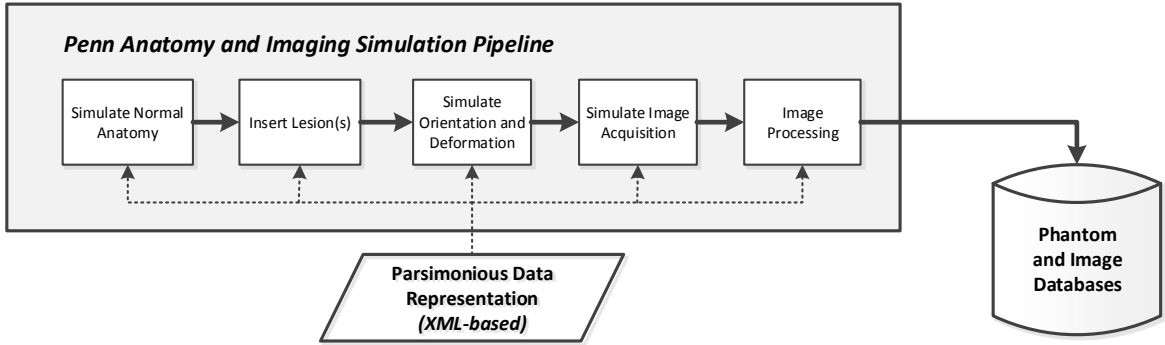

**Fig. 1**. Schematic of the breast simulation and image acquisition pipeline. The pipeline executes VCT based on an XML data file. Simulated images are archived for the virtual reader studies.

The display of the images is simulated using the MeVIC simulation pipeline [22]. A block diagram of the proposed observer model is given in Fig. 2. The blocks within the dotted rectangle are introduced in this paper and described in the remainder of this section. The rest of the blocks have been explained previously [1, 2, 4].

A brief description of the proposed method follows. After simulation of the display and HVS, a number of features indicative of lesions presence in the stack are calculated. The estimated level of background complexity (roughly corresponding to how many lesion-like objects are seen in the stack) is used to adjust the level of confidence in features by modulating the amount of Gaussian noise added to the feature values (The higher the estimated level of background complexity, the lower is the confidence in lesion features). Stacks are ranked based on each feature and these results are combined into a single rank for each stack that is used in the ROC analysis and AUC (area under ROC curve) calculation.

To create datasets with varying background complexity, spatiotemporal low-pass Gaussian noise with four different levels of energy was added to the dataset generated by the UPenn phantoms [4].

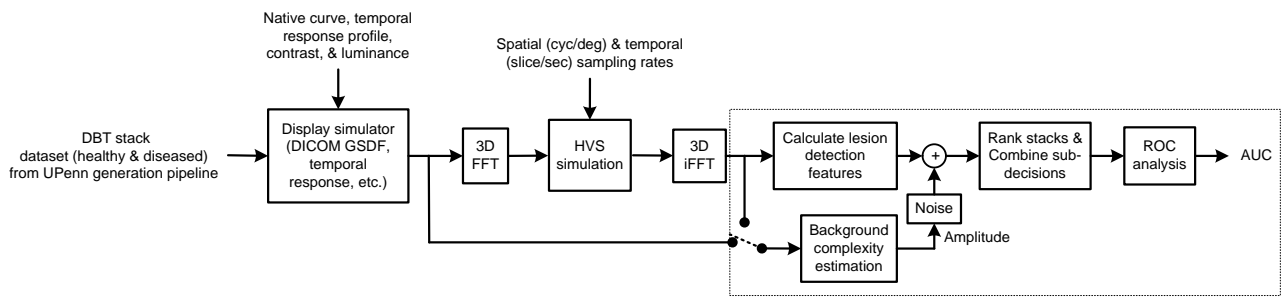

**Fig. 2.** Block diagram of the proposed anthropomorphic virtual observer. Blocks within the dotted rectangle are the focus of this paper. Background complexity may be estimated from before- (Section 3.1.2) or after-HVS (Section 3.1) simulation stacks.

## 2.1 Human reader error analysis

Among the six varieties of the anthropomorphic model observer that we have proposed [4], "PM+masking" caused the largest change in AUC (from 97% to 92%; d' from 2.66 to 1.99)[*] when background complexity changed from level 0 (simple) to 4 (very complex; several lesion-like objects visible). For the same change in background complexity, the performance of human observers dropped considerably more (from AUC = 98.6% to 67.5% and d'=3.11 to 0.64 for observer A; from 92.6% to 60.6% and d' = 2.05 to 0.38 for observer B) [4].

In the reader study referenced above, from the same datasets used for model observer simulations, 35 DBT stacks (each consisting of 32 slices with 64x64 pixels) were randomly chosen for each of the six conditions. A total of six conditions were examined {lesion, healthy} for each of 3 levels of background complexity. Note that the lesion, if present, is always in the spatiotemporal center of the image stack (i.e., a location known exactly). Scores {0, 1, 2, 3} meant {certain lesion absent, probably lesion absent, probably lesion present, certain lesion present}. Histograms of the scores of an observer assigned to lesion ("signal present") and healthy ("signal absent") cases in the lowest (0) and highest (4) background complexity levels are depicted in Fig. 3. Behavior of the second observer followed a similar trend. It may be concluded that the errors made by human observers in highly complex background are more commonly false negatives (missing lesions) than false positives (detecting erroneous lesions due to noise).That is because in high complexity background, the histogram for signal present cases became similar to the histogram for signal absent cases (i.e., heavier on low scores).

Five examples of lesions missed by human observers in a complex background are shown in Fig. 4. In each case, the lesion slice as well as its difference with the slice before and after are shown to demonstrate the "pop-out" effect. In the pop-out effect, a bright spot in the middle of an image (i.e., a lesion) is visible because the proceeding and succeeding slices are of lower brightness and the dark-bright-dark sequence causes the lesion to flash or pop-out to the observer. The differences shown in Fig. 4 are calculated such that a bright spot in the middle of each slice is an indication of a lesion. It seems that to receive a high score (i.e., lesion present) from the human observer, the presence of multiple clues is necessary. For example, in cases 1 and 4 shown in Fig. 4, the lesion is visibly brighter than the region of interest (ROI) in the slice after, but it is not brighter than its surround in its own slice nor is brighter than the ROI in the slice before. In case 3, the lesion is brighter than ROI in the slice before and after, but not brighter than its surround. In essence, the lesion is appended to the top or bottom surface of a normal structure of similar signal intensity. As a result, the object appears to a human observer as being slightly elongated, rather than being seen as a lesion separate from the background.

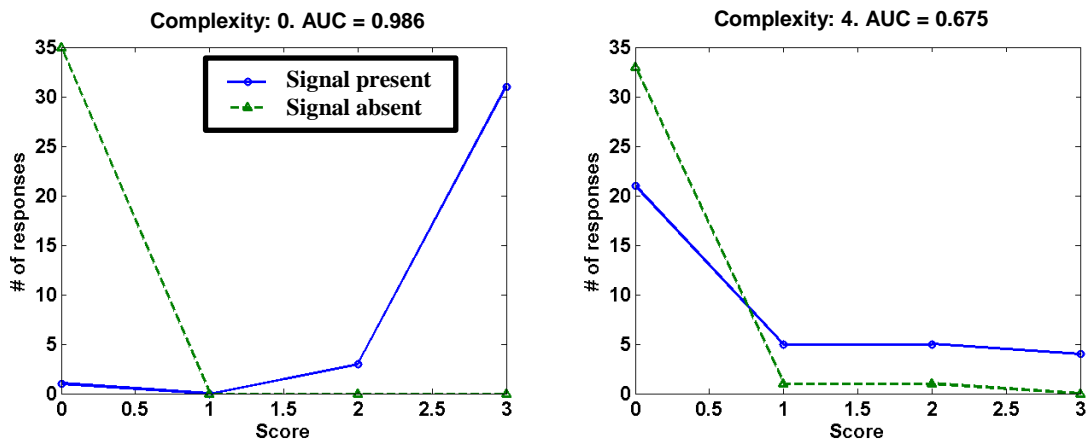

**Fig. 3.** Distribution of scores given by human observer A in simple (left) and highly complex (right) background.

---

[*] Detectability index, d', is calculated from AUC using d' = 2 erf$^{-1}$(2 AUC – 1), where erf$^{-1}$( ) is the inverse error function.

## 2.2 Derivation of anthropomorphic "channels" (visual cues for lesion)

Based on these observations, we theorize that the human observer's decision about the presence of a lesion can be modeled as logical AND of the results of several sub-decisions, each determining the presence of a visual cue for the lesion. This is in contrast to a typical model observer in which the detection is based on thresholding a single score. The concept of performing independent detection of parts/components/cues and then making a decision by pooling the sub-decisions has been proposed before; for example, Pelli *et al* [5] has proposed this as a mechanism of how humans recognize letters of alphabet.

Based on the analysis above, we propose the following anthropomorphic channels (i.e., visual cues or features):

$f_1$: "How much brighter is the lesion ROI than its surround in the same slice?" This is calculated as the difference between the average signal of the lesion (a 3-pixel radius disk in the center of slice 16) and the lesion surround (a ring with inner and outer radii of 5 and 7 about the center of slice 16);

$f_2$: "How much brighter is the lesion ROI than the prior slice?" This is calculated as the difference between the average signal of the lesion (a 3-pixel radius disk in the center of slice 16) and the lesion ROI in the slice prior (a 4-pixel radius disk in the center of slice 15);

$f_3$: "How much brighter is the lesion ROI than the subsequent slice?" This is calculated as the difference between the average signal of the lesion (a 3-pixel radius disk in the center of slice 16) and the lesion ROI in the slice after (a 4-pixel radius disk in the center of slice 17);

$f_4$: "What is the confidence in the above brightness comparisons?" Several candidate measures of confidence are detailed in Section 2.3.

Note that these are simple cues that we only propose for the detection of a small target embedded in a single slice. For example, in detecting a mass, a human reader probably uses features related to the edges of the mass. Similarly, a mass typically spans many slices, thus the human observer will incorporate information from more than three slices. Thus, the sub-decision criteria should be adjusted according to the detection or classification task.

Note that the confidence in brightness comparison is inversely related to the background complexity as explained below. Consider the first feature above, $f_1$, as an example. Let $I_{lesion} \pm \sigma_1$ and $I_{surround} \pm \sigma_2$ denote the quantities being compared, where $\sigma_i$ is the standard deviation in measuring the brightness. Thus, we have $f_1 = (I_{lesion} - I_{surround}) \pm \sigma$, where $\sigma^2 = \sigma_1^2 + \sigma_2^2$, assuming independent measurement of lesion and surround brightness. The quantity $\sigma^2$ is positively correlated with the local energy of the image and the background complexity. Thus, a higher background complexity causes a higher $\sigma$, which in turn means higher uncertainty (deviation) in value of $f_1$. This example is used to show that the validity of a brightness comparison is a function of background complexity, but it cannot answer what measure of confidence a human observer uses.

## 2.3 Confidence in lesion detection

As pointed out in Section 2.1, human readers perform worse for the lesion detection task in DBT stacks with a complex background not because of false positives but because of false negatives. To include the adverse effect of background complexity on lesion detection in our model observer, we investigated methods for automatic compensation of the background complexity.

Mainprize *et al* proposed a local signal to noise ratio, $d_{local}$ defined below, as a metric of the apparent density in mammograms, hence a metric for lesion masking possibility (Eq. (1) in [6]):

$$d_{local}^2 = \frac{(\int T^2(u,v) W^2(u,v) \, du \, dv)^2}{\int NNPS^2(u,v) T^2(u,v) W^2(u,v) \, du \, dv}$$

where $T$ is the modulation transfer function, $W$ is the task function, *NNPS* is the normalized "noise" power spectrum (noise includes an anatomical component due to breast structure), and $(u,v)$ are spatial frequencies. $d_{local}$ is calculated for each ROI. Direct calculation of the *NNPS* for each ROI is not possible. Therefore, a model of *NNPS* [10] is calibrated using the noise measured for each ROI. Considering that the numerator of $d_{local}$ expression (above) is independent of the ROI, for a given mammogram and a given system, $d_{local}$ is inversely proportional to ROI energy (local energy of the image). Therefore, we considered the energy of the lesion ROI as well as the energy of the whole slice (#16 which has the lesion in positive stacks) as candidate features, denoted by $b_1$ and $b_2$ respectively, indicating the level of background complexity.

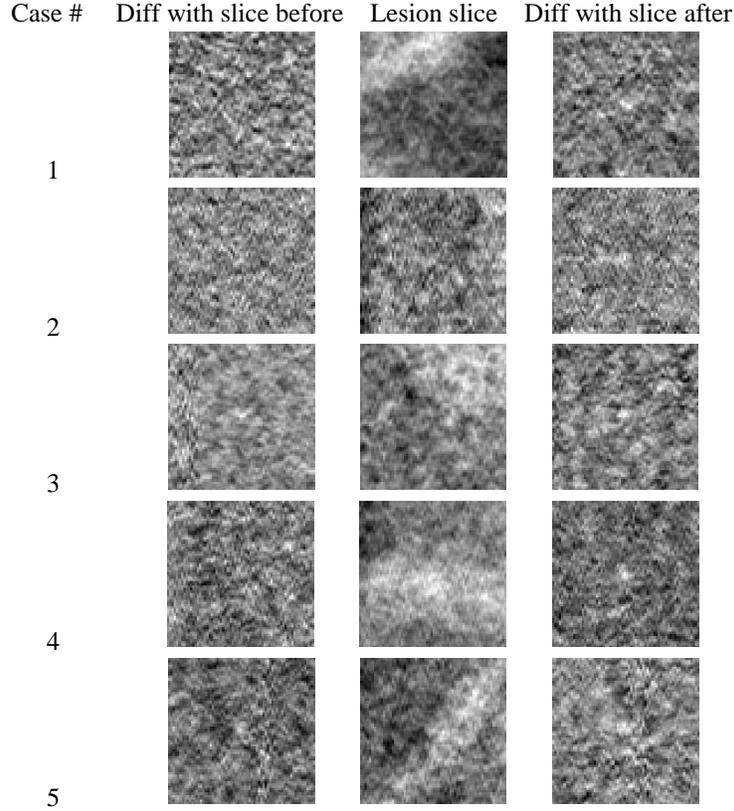

**Fig. 4.** Each row is a lesion case, with level 4 background complexity, missed by human observer A. Score 0 was assigned to the top three cases and score 1 was assigned to the bottom two. *Middle column*: the slice with the lesion in the center. *Left column*: the slice before subtracted from lesion slice. *Right column*: the slice after subtracted from lesion slice.

Since the stacks are displayed in cine mode and the location of the lesion (if present) is known in our reading setup, we conjectured that another reason for the reduced performance in complex backgrounds is perhaps due to the extraneous lesion-like objects that pop out more when browsing stacks with higher background complexity. There will be more such objects when consecutive slices differ more. This motivated us to consider the difference between consecutive slices as a feature of background complexity. To calculate the difference between consecutive slices we used peak signal-to-noise ratio (PSNR) [13, 14] and structural similarity index (SSIM) [11, 12]. PSNR quantifies the amount of difference between two images, *X* and *Y*, and is defined as

$$PSNR = 20\log_{10}\max(X) - 10\log_{10}\overline{(X-Y)^2}$$

where max(*X*) is the maximum possible value of the image (e.g., 255 for 8-bit images), and *X-Y* is the pixel-wise subtraction and the bar indicates averaging over all pixels. Thus, the more similar *X* and *Y* (e.g., low background complexity, causing consecutive slices to be more similar in our case), the higher is PSNR value. SSIM also quantifies the difference between two images; however, it is known to be correlated to the perceived similarity between two images (DBT slices in this case). Accordingly, we define the following features

$$b_3 = \{p_1, p_2, \ldots, p_{31}\}, \text{ where } p_i = PSNR\{s_i, s_{i+1}\} ,$$

$$b_4 = \{q_1, q_2, \ldots, q_{31}\}, \text{ where } q_i = SSIM\{s_i, s_{i+1}\} ;$$

here, $s_i$ denotes the $i^{th}$ slice of a stack.

The potential of features $b_1$ to $b_4$ in estimation of background complexity level is discussed Sections 2.5 and 3.1. Once background complexity level is estimated, zero-mean Gaussian noise modulated with complexity level is added to the features $f_1$ to $f_3$ to deteriorate detection performance more with higher background complexity.

### 2.4 Combining sub-decisions

After considering the effect of background complexity as described above, we use $f_1$ to $f_3$ to rank the DBT stacks as described below. The process is similar to the Wilcoxon method of ROC analysis, since the stacks are ranked based on their scores. However, since we have three scores (i.e., values of features $f_1$ to $f_3$) per stack, we will perform three sort operations, which yield three rankings for stacks. Thus, depending on the feature values, three (possibly different) ranks are assigned to each stack. To conduct the AND operation between the sub-decisions, for each stack, we take the minimum of the three ranks as the stack rank and use those as the input scores for Wilcoxon AUC calculation method, which can take a single score per stack. Using the minimum of ranks based on each of the features provides an effect similar to what observed in Fig. 3: when ranks from all features are the same, the result will have the same distribution as each of the ranks, similar to signal present distribution in simple background (Fig. 3, left diagram). When ranks are a bit different (e.g., due to high background complexity distorting $f_1$ to $f_3$ values), the result will have a distribution heavier on lower ranks, similar to signal distribution in high background complexity (Fig. 3, right diagram). This effect is explained with an example (Fig. 5).

| Features agree (simple background) | Features disagree (complex background) |
|---|---|
| $f_1$: 1, 2, 3, 4, 5 | 2, 1, 3, 4, 5 |
| $f_2$: 1, 2, 3, 4, 5 | 2, 3, 1, 4, 5 |
| $f_3$: 1, 2, 3, 4, 5 | 1, 3, 2, 4, 5 |
| Min. rank: 1, 2, 3, 4, 5 | 1, 1, 1, 4, 5 |

**Fig. 5.** Illustration of how minimum rank combination of sub-decisions can simulate the effect of background complexity observed in distribution of human reader scores under simple and complex background. Higher rank means stronger lesion. When sub-decisions disagree (slightly different ranking based on individual features; right column), distribution of minimum rank results is heavier in lower ranks. Note that the minimum rank values are considered the scores used as input of ROC analysis.

### 2.5 Measuring power of features in identifying background complexity level

The performance of a Hotelling observer was used to measure the effectiveness of various sets of features, $F_{n \times 1}$, with $n > 1$, in the background complexity estimation task. The observer, $w$, is calculated using

$$w_{n \times 1} = (\Sigma_0 + \Sigma_1)^{-1} (\mu_1 - \mu_0),$$

where $\Sigma_0$ and $\Sigma_1$ are the covariance matrixes of features for the training subset of simple and highly complex background stacks respectively. Mean of features for such stacks are shown with n x 1 vectors $\mu_0$ and $\mu_1$. For n = 1, i.e., a single feature set, this result is simplified to w = 1.

The performance of this observer, in terms of AUC, in classifying the stacks in the test subset with scores $w^T F_1$ (for positive stacks) and $w^T F_0$ (for negative stacks) is used to measure the power of features in identifying background complexity. Positive (negative) means high complexity (simple) background. Hence, the maximum power, resulting in perfect separation of simple and highly complex backgrounds will be 1, and the minimum power (equivalent to picking by chance) will be 0.5.

We used all 2632 image stacks, for both training as well as testing. Half of the stacks are healthy (lesion absent) and half contain a lesion; similarly, half have a simple background and half have a complex background. Doing so is likely to overestimate the power of features but this estimate is adequate for comparative purposes.

# 3. RESULTS

## 3.1 Automatic estimation of background complexity level

Using the method of Section 2.5, we compare the power of features $f_1$ to $f_3$ (Section 2.2) and $f_4$ given by $b_1$ to $b_4$ (Section 2.3) in measuring the level of background complexity. The results are listed in Table 1. As compared to $b_4$, feature set $b_3$ provides the same distinction in background complexity level, at about 40 times less computational cost.

**Table 1.** A comparison between candidate feature sets in terms of their power in identifying the level of background complexity. Features are defined in Section 2.2 and Section 2.3.

| Feature set | Power (in AUC) | Power (in d') |
|---|---|---|
| $\{f_1, f_2, f_3\}$ | 0.517 | 0.06 |
| $f_4 = b_1$ | 0.505† | 0.02 |
| $f_4 = b_2$ | 0.510 | 0.04 |
| $f_4 = b_3$ | 0.822 | 1.31 |
| $f_4 = b_4$ | 0.813 | 1.26 |

### 3.1.1 Reducing dimensionality of features

We inspected the Hotelling observer coefficients, $w$ defined in Section 2.5, to explore possibility of computational savings by reducing the dimensionality of $b_3$ (Fig. 6). By picking the five largest magnitude coefficients (i.e., five slice differences that are most influential in background complexity level) in Fig. 6, we construct $b'_3 = \{p_2, p_3, p_{27}, p_{28}, p_{30}\}$, as a candidate feature set, a subset of $b_3$, for estimation of background complexity level, with $p_i$ defined in Section 2.3. As compared to $b_3$, feature set $b'_3$ is about 6 times computationally less expensive to calculate maintaining a good power (AUC = 0.799). However, note that the makeup of $b'_3$ can change as a function of the training subset. That is because $b_3$, and thus it's largest magnitude coefficients, are calculated from the training subset.

---

† This AUC value was originally 0.495. By swapping "signal" and "noise" scores, it increases to 0.505.

### 3.1.2 Background complexity estimation from stacks before HVS simulation

By HVS simulation, some of the visual information in a sequence is lost. That is because HVS model roughly acts as a band-pass spatiotemporal filter, attenuating low and high frequency components (thus, reducing information) available in the before-HVS stack. In fact, that is why observer models without HVS simulation consistently outperform human readers. Better estimation of background complexity level is possible if stacks before the HVS simulation are used. For example, power of $b_3$ when calculated from before-HVS stacks is increased from AUC = 0.822 to 0.99955 (d' is increased from 1.3 to 4.7).

### 3.2 Detecting lesions with background complexity features

It is always desirable to have fewer features extracting the required information from the data. We noted that lesion detection features $f_1$ to $f_3$ (Section 2.2) do not convey information regarding background complexity level (Table 1). We also investigated if $b_3$ or $b_4$ can be used for detecting lesions as well as background complexity level. To that end, we adapted the method of Section 2.5 for lesion detection task (i.e., measured power of features for lesion detection task):

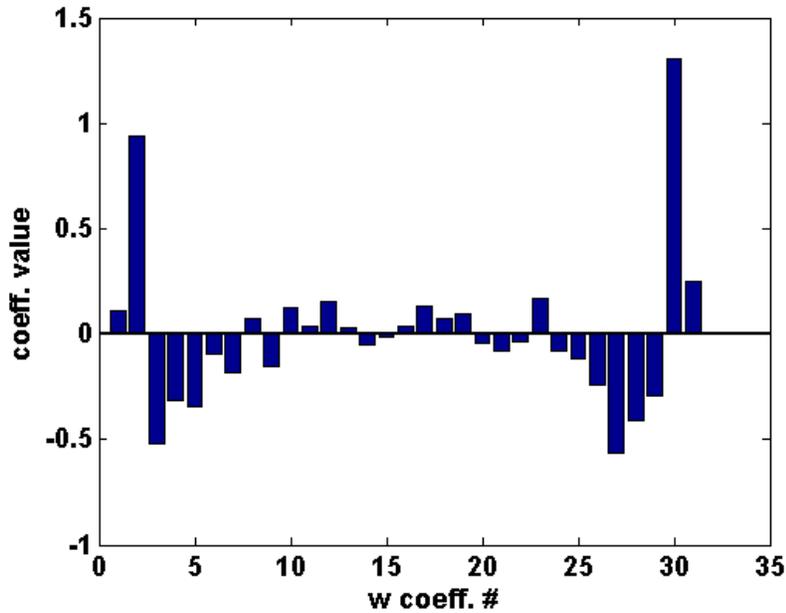

**Fig. 6.** Weight of each of the slice differences in determining the background complexity level.

Instead of simple and highly complex background stacks, the observer is trained to discriminate between healthy and lesion stacks. The power of feature sets $b_3$ and $b_4$ calculated this way are rather low: 0.572 and 0.566. Hence, we must rely on $f_1$ to $f_3$ for lesion detection. Therefore, we conclude that the minimum set of features we need to design a model observer that tracks human reader performance in various background complexity levels must include both lesion detection features $f_1$ to $f_3$, as well as background complexity features $b_3$ or $b_4$.

### 3.3 Overall results

Lesion detection performance in simple and highly complex background, using different background complexity estimation methods, is calculated via simulation. To estimate AUC variations, the data is partitioned into four groups of equal size. Three of the groups were used to train three instances of background complexity estimator. The last group was used to calculate the lesion detection performance with each of the three estimators. Average AUC and 95% confidence interval (i.e., ± twice standard deviation; when available) are listed in Table 2 for background estimation ideal, before-HVS, after-HVS and none. Histograms of scores, the input to ROC analysis, for signal present (lesion) and signal absent (no lesion) stacks, in simple and highly complex background conditions, are provided in Fig. 7. Results of human and numerical observers reported in [4] are listed in Table 3 for comparison. Note that human observers only read a small subset of stacks or the reading session would be impractical and tiring.

**Table 2.** Comparison of lesion detection performance in terms of AUC using different background complexity estimation methods.

| **Background complexity level ↓** | **Background complexity estimation method** | | | |
|---|---|---|---|---|
| | After-HVS b3 | Before-HVS b3 | Ideal background complexity estimation | No background complexity estimation |
| Simple | 0.730±0.039 | 0.782±0.027 | 0.790 | 0.790 |
| Highly complex | 0.576±0.029 | 0.509±0.002 | 0.504 | 0.754 |

Note that with the lesion detection features $f_1$ to $f_3$ (Section 2.2) and minimum rank combination of sub-decisions (Section 2.4), AUC cannot be further increased for simple background, as no extra noise is added to the features at this complexity level.

**Table 3.** Lesion detection performance in terms of AUC for human readers and a select model observer from [4].

| **Background complexity level ↓** | **Observer** | | |
|---|---|---|---|
| | PM+masking Model observer [4] | Human reader A [4] | Human reader B [4] |
| Simple | 0.97 | 0.986 | 0.926 |
| Highly complex | 0.92 | 0.675 | 0.606 |

For further comparison, results of the proposed model observer with before-HVS background complexity estimation and results from [4] are shown in Table 4 in terms of d'.

**Table 4.** Lesion detection performance in terms of d' for human readers, a select model observer from [4], and the proposed model observer with before-HVS background complexity estimation.

| **Background complexity level ↓** | **Observer** | | | |
|---|---|---|---|---|
| | PM+masking Model observer [4] | Human reader A [4] | Human reader B [4] | Proposed method with before-HVS background complexity estimation |
| Simple | 2.66 | 3.11 | 2.05 | 1.1 |
| Highly complex | 1.99 | 0.64 | 0.38 | 0.03 |
| Drop in performance (difference of rows above) | **25.2%** | **79.4%** | **81.5%** | **97.3%** |

## 4. DISCUSSION

One question that may arise is why can we not simulate human observer performance trends with background complexity even though we are modeling the properties of HVS? The answer, we believe, lies in the fact that the properties of HVS modeled in our observer so far govern low-level vision and are developed based on the results of simple experiments (e.g., detection of 1-D tonal spatial or spatiotemporal patterns). Most likely higher levels of cognition are involved in a task as complicated as the detection of lesion in a complex background.

The ideal set of features and/or method of combining the sub-decisions would produce score histograms that match those of human readers in simple and highly complex backgrounds. From this perspective, the features we used or the minimum rank combination of sub-decisions are not ideal (cf. Fig. 7 and Fig. 3). To close this gap, the feature set should be more powerful in detecting lesions in simple background (more distinct signal present and signal absent distributions in left columns) and yield the same signal absent distribution in complex background. This is unlike, for example, the signal absent distribution using after-HVS background estimation shown in Fig. 7 middle graphs, where the signal absent histogram demonstrates a shift in peak from simple to highly complex background conditions.

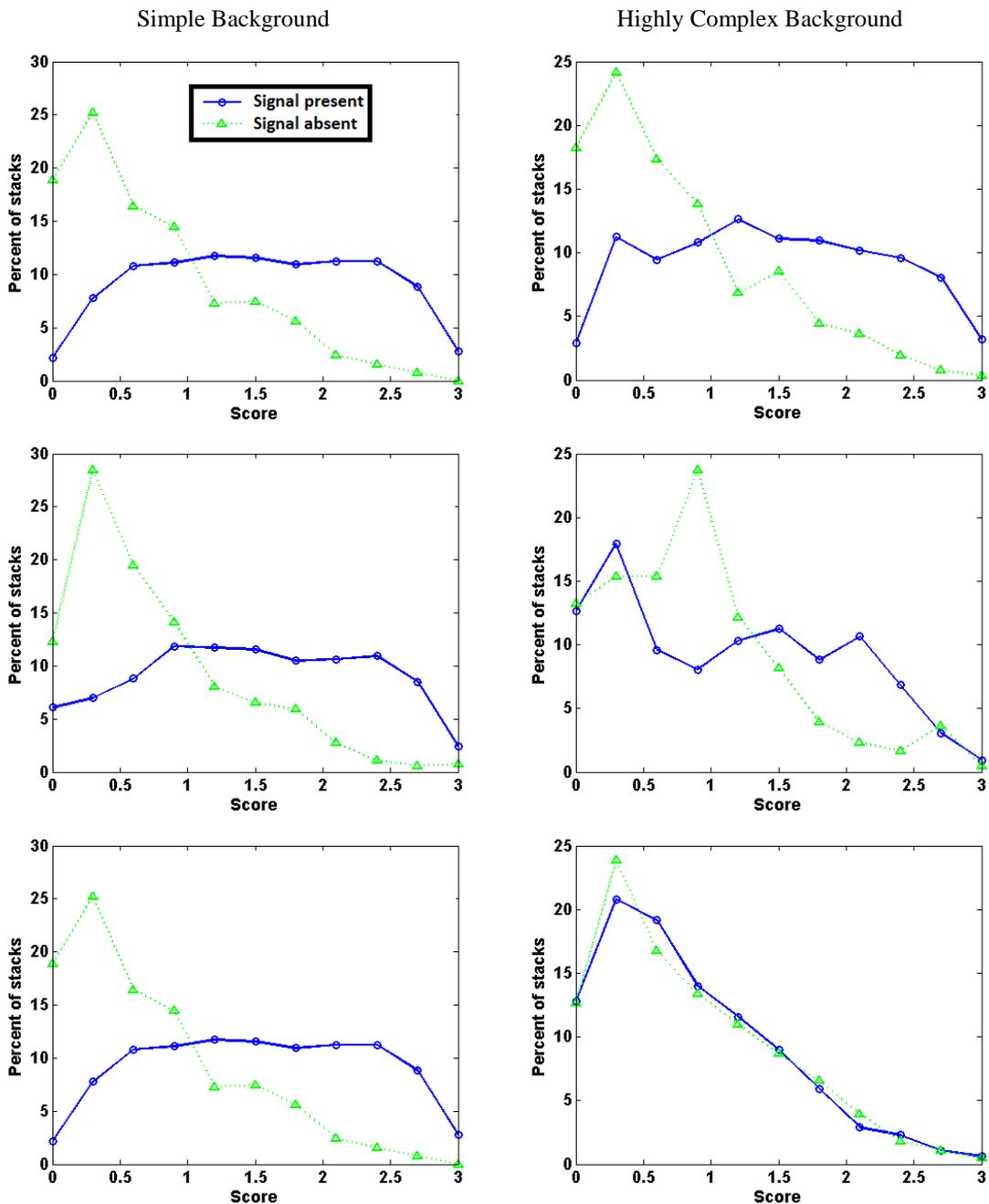

**Fig. 7.** Histogram of scores for stacks with simple (left column) and highly complex (right column) backgrounds, using no background estimation (top row), after-HVS background estimation (middle row) and ideal background estimation (bottom row). The latter is the same for before-HVS background estimation. Scores are scaled to match horizontal axis of Fig. 3.

# 5. CONCLUSION

Based on the results listed in Table 4, we conclude that the proposed model observer has a larger drop in detection performance between simple and highly complex background conditions as compared to the previous work. Our results demonstrate that the performance of the proposed model observer degrades substantially with background complexity similar to human observers. There is room for improvement in the degree of degradation.

Our research in development of an anthropomorphic observer is a work in progress. While showing promise, it has certain limitations and further work is required as described below.

One limitation of this work is that the set of lesion detection features we considered ($f_1$ to $f_3$; Section 2.2) is incomplete. This may be deduced from the fact that the detection performance of the proposed model observer is less than human readers in a simple background (cf. Table 2 and 3), or from the fact that signal-present histograms for simple background (Fig. 7 left column) do not peak at maximum score, unlike a human reader histogram (Fig. 3 left graph). An analysis of detection failure cases (similar to Section 2.1) may be used to derive features that complement the existing lesion detection feature set.

Our study is also limited to the type of lesion (i.e., single micro-calcification) considered. For detection of a more complex lesion (e.g., spiculated mass), not only we need a different set of anthropomorphic features for lesion detection to mimic human reader's detection of the complex lesion, but also we may need a more sophisticated way to combine sub-decisions (made based on subsets of features) such as a decision flowchart instead of the minimum rank combination. Moreover, the background complexity for a complex lesion must be redefined and its estimation method reworked. That is because it may be less likely for noise lumps to constitute something that can be possibly mistaken with the sought after complex lesion.

Future directions of this work include finding a more robust set of anthropomorphic features (that can be used for lesions other than micro-calcifications, and/or yields score distribution similar to human readers) and a better way to combine the sub-decisions.

# ACKNOWLEDGEMENT

Ali Avanaki would like to thank Eddie Knippel, Albert Xthona, and Cédric Marchessoux.

# REFERENCES

[1] Avanaki, A. R. N., Espig, K. S., Maidment, A. D. A., Marchessoux, C., Bakic, P. R., & Kimpe, T. R. L. (2014, March). Development and evaluation of a 3D model observer with nonlinear spatiotemporal contrast sensitivity. In *SPIE Medical Imaging* (pp. 90370X-90370X). International Society for Optics and Photonics.

[2] Avanaki, A. R. N., Espig, K. S., Marchessoux, C., Krupinski, E. A., Bakic, P. R., Kimpe, T. R. L., & Maidment, A. D. A. (2013, March). Integration of spatio-temporal contrast sensitivity with a multi-slice channelized Hotelling observer. In *SPIE Medical Imaging* (pp. 86730H-86730H). International Society for Optics and Photonics.

[3] Platiša, L., Goossens, B., Vansteenkiste, E., Park, S., Gallas, B. D., Badano, A., & Philips, W. (2011). Channelized Hotelling observers for the assessment of volumetric imaging data sets. *JOSA A*, *28*(6), 1145-1163.

[4] Avanaki, A. R. N., Espig, K. S., Xthona, A., Kimpe, T. R. L., Bakic, P. R., & Maidment, A. D. A. (2014). It is hard to see a needle in a haystack: Modeling contrast masking effect in a numerical observer. In *Breast Imaging* (pp. 723-730). Springer International Publishing.

[5] Pelli, D. G., Farell, B., & Moore, D. C. (2003). The remarkable inefficiency of word recognition. *Nature*, *423*(6941), 752-756.

[6] Mainprize, J. G., Wang, X., Ge, M., & Yaffe, M. J. (2014). Towards a Quantitative Measure of Radiographic Masking by Dense Tissue in Mammography. In *Breast Imaging* (pp. 181-186). Springer International Publishing.

[7] Pokrajac, D. D., Maidment, A. D. A., & Bakic, P. R. (2012). Optimized generation of high resolution breast anthropomorphic software phantoms. *Medical physics*, *39*(4), 2290-2302.

[8] Lago, M. A., Maidment, A. D. A., & Bakic, P. R. (2013). Modelling of mammographic compression of anthropomorphic software breast phantom using FEBio. In *Proc. Int'l Symposium on Computer Methods in Biomechanics and Biomedical Engineering (CMBBE). Salt Lake City, UT*.

[9] Kuo, J., Ringer, P. A., Fallows, S. G., Bakic, P. R., Maidment, A. D., & Ng, S. (2011, March). Dynamic reconstruction and rendering of 3D tomosynthesis images. In *SPIE Medical Imaging* (pp. 796116-796116). International Society for Optics and Photonics.

[10] Mainprize, J. G., & Yaffe, M. J. (2010). Cascaded analysis of signal and noise propagation through a heterogeneous breast model. *Medical physics*, *37*(10), 5243-5250.

[11] Wang, Z., Bovik, A. C., Sheikh, H. R., & Simoncelli, E. P. (2004). Image quality assessment: from error visibility to structural similarity. *Image Processing, IEEE Transactions on*, *13*(4), 600-612.

[12] http://en.wikipedia.org/wiki/Structural_similarity, Accessed Feb 2015.

[13] Huynh-Thu, Q., & Ghanbari, M. (2008). Scope of validity of PSNR in image/video quality assessment. *Electronics letters*, *44*(13), 800-801.

[14] http://en.wikipedia.org/wiki/Peak_signal-to-noise_ratio, Accessed Feb 2015.

[15] Barten, P. G. (1999). *Contrast sensitivity of the human eye and its effects on image quality*. SPIE press.

[16] Bakic PR, Albert M, Brzakovic D, Maidment ADA. Mammogram synthesis using a 3D simulation. II. Evaluation of synthetic mammogram texture. *Medical Physics*. 2002;29(9):2140-51.

[17] Bakic PR, Albert M, Brzakovic D, Maidment ADA. Mammogram synthesis using a 3D simulation. I. Breast tissue model and image acquisition simulation. *Medical Physics*. 2002;29(9):2131-9.

[18] Bakic PR, Albert M, Brzakovic D, Maidment ADA. Mammogram synthesis using a three-dimensional simulation. III. Modeling and evaluation of the breast ductal network. *Medical Physics*. 2003;30(7):1914-25.

[19] Carton AK, Bakic P, Ullberg C, Derand H, Maidment AD. Development of a physical 3D anthropomorphic breast phantom. *Med Phys*. 2011 Feb;38(2):891-6.

[20] Shankla V, Pokrajac D, Weinstein SP, DeLeo M, Tuite C, Roth R, et al. Automatic insertion of simulated microcalcification clusters in a software breast phantom. *SPIE Medical Imaging 2014: Physics of Medical Imaging*; San Diego, CA: SPIE, 9033, p., 2014.

[21] Maidment ADA, Bakic PR, Ruiter NV, Richard FJP. Model-based comparison of two breast tissue compression methodologies. *Medical Physics.* 2004 Jun;31(6):1786-.

[22] Marchessoux, C., Kimpe, T. R. L., & Bert, T. (2008). A virtual image chain for perceived and clinical image quality of medical display. *Display Technology, Journal of*, *4*(4), 356-368.